\pdfoutput=1
\documentclass[runningheads]{llncs}

\usepackage{eccv}
\usepackage{eccvabbrv}
\usepackage{graphicx}
\usepackage{wrapfig}
\usepackage{booktabs}
\usepackage{makecell}
\usepackage{siunitx}
\usepackage{amsmath,amssymb}
\usepackage[accsupp]{axessibility}
\usepackage{hyperref}

\hypersetup{
  pdftitle={AGORA: Adversarial Generation Of Real-time Animatable 3D Gaussian Head Avatars},
  pdfauthor={Ramazan Fazylov, Sergey Zagoruyko, Aleksandr Parkin, Stamatis Lefkimmiatis, Ivan Laptev}
}

\begin{document}

\title{\texorpdfstring{AGORA: \underline{A}dversarial \underline{G}eneration \underline{O}f \underline{R}eal-time \underline{A}nimatable 3D Gaussian Head Avatars}{AGORA: Adversarial Generation Of Real-time Animatable 3D Gaussian Head Avatars}}
\titlerunning{AGORA: Animatable 3D Gaussian Head Avatars}

\author{Ramazan Fazylov\inst{1} \and Sergey Zagoruyko\inst{2} \and Aleksandr Parkin\inst{3} \and Stamatis Lefkimmiatis\inst{3} \and Ivan Laptev\inst{1}}
\authorrunning{R. Fazylov et al.}
\institute{Mohamed bin Zayed University of Artificial Intelligence, Abu Dhabi, UAE
\and
Polynome AI, Dubai, UAE
\and
MTS AI, Moscow, Russia
}

\maketitle

\begin{abstract}
The generation of high-fidelity, animatable 3D human avatars remains a core challenge in computer graphics and vision, with applications in VR, telepresence, and entertainment. Existing approaches based on implicit representations like NeRFs suffer from slow rendering and dynamic inconsistencies, while 3D Gaussian Splatting (3DGS) methods are typically limited to static head generation, lacking dynamic control. We bridge this gap by introducing AGORA, a novel framework that extends 3DGS within a generative adversarial network to produce animatable avatars. Our key contribution is a lightweight, FLAME-conditioned deformation branch that predicts per-Gaussian residuals, enabling identity-preserving, fine-grained expression control while allowing real-time inference. Identity is further preserved through spatial shape conditioning of the identity branch, and expression fidelity is enforced via a dual-discriminator training scheme leveraging synthetic renderings of the parametric mesh. AGORA generates avatars that are not only visually realistic but also precisely controllable. Quantitatively, we outperform state-of-the-art NeRF-based methods on expression accuracy while rendering at 250 FPS on a single GPU and, notably, at $\sim$9 FPS under CPU-only inference -- to our knowledge the first demonstration of CPU-only animatable 3DGS avatar synthesis. This work represents a significant step toward practical, high-performance digital humans.
\end{abstract}

\vspace{-.2cm}\section{Introduction}
\label{sec:intro}

The creation of realistic and controllable digital humans is a significant area of research within computer graphics and computer vision. Demand for high-fidelity avatars is rapidly increasing across applications ranging from immersive VR/AR experiences and telepresence to the entertainment industry. For these applications to be effective, avatars must not only look realistic but also be able to produce accurate expressions. However, generating avatars that are visually convincing, easily animatable, and computationally efficient remains a complex challenge that motivates the exploration of new synthesis and animation techniques.

Recent progress in avatar creation can be broadly categorized into three paradigms. The first, reconstruction-based methods, optimize a 3D representation to fit multi-view \cite{GaussianAvatars,GHA,NPGA, Wang2025CVPR} or monocular videos \cite{nha,nerface,insta,pointavatar, imavatar, monogausavatar}. While capable of producing high-fidelity results for a specific subject, these approaches require lengthy per-subject optimization and often struggle to generalize to novel expressions. 
A second paradigm leverages large-scale 2D diffusion models via score distillation sampling to generate 3D assets from text \begin{wrapfigure}{r}{0.50\textwidth}
    \centering
    \vspace{-10pt}
    \includegraphics[width=\linewidth]{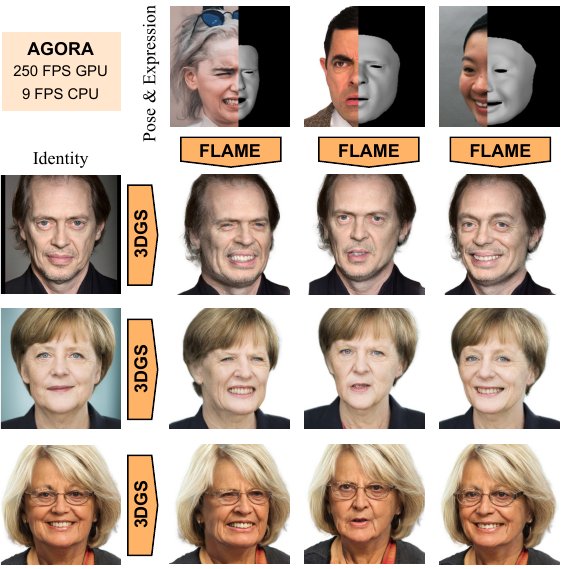}
    \caption{Our method, AGORA, generates animatable 3D avatars. AGORA produces highly photorealistic identity-preserving results and supports dynamic control on pose and face expressions, while allowing real-time inference at 250 FPS on a single GPU and $\sim$9 FPS under CPU-only inference. }
    \label{fig:teaser}
    \vspace{-14pt}
\end{wrapfigure} (SDS) \cite{DreamFusion, HeadSculpt, HeadStudio, GAF} or a single image. However, these methods often face challenges with view-consistency and achieving the level of detail required for realistic human heads. The most promising direction for generating diverse, high-quality avatars has been 3D-aware Generative Adversarial Networks (3D GANs) \cite{eg3d, gghead, gsgan, panohead, spherehead, epigraf, voxgraf, giraffe,  gram, pigan, graf, gdna}, which learn a distribution of 3D heads from 2D image collections \cite{stylegan, VFHQ}. While some works have extended 3D GANs to dynamic scenarios \cite{Next3D, DiscoFaceGAN, GNARF, anifacegan, OmniAvatar, fenerf, eva3d}, they typically rely on computationally expensive neural fields. Concurrently, recent advances in static generation have adopted the highly efficient 3D Gaussian Splatting representation \cite{gsgan, gghead}, but lack animation capabilities. This leaves a critical gap: a method that combines the generative power of 3D GANs with the real-time rendering of 3DGS to create fast, explicit and precisely-animatable avatars. Recent work, GAIA \cite{gaia}, attempts to fill this gap by incorporating expression conditioning to produce animatable avatars. However, this method, while promising, still faces challenges in both inference speed and visual fidelity.

To address this, we introduce a novel framework for generating animatable 3D head avatars that unites the efficiency of 3DGS with the generative power of 3D GANs. Our approach builds upon a static generator, inspired by GGHEAD \cite{gghead}, which produces a canonical set of 3DGS attributes. To enable high-fidelity control, we introduce spatial shape conditioning in the generator and a lightweight, FLAME-conditioned \cite{FLAME} deformation branch that predicts attribute residuals for canonical Gaussians. We further enforce expression faithfulness with a dual-discriminator scheme that provides adversarial supervision on both rendered appearance and synthetic geometry cues. Since the identity-specific generator has to be evaluated only once per subject, animation reduces to the lightweight deformation branch, which keeps the whole pipeline real-time.
Crucially, we train from static face images only, without multi-view capture or video supervision.
The resulting model generates high-fidelity, controllable avatars capable of real-time rendering (Fig.~\ref{fig:teaser}).
Our contributions can be outlined as follows:
\vspace{-.2cm}
\begin{itemize}
    \item A novel animatable 3DGS generator that couples a lightweight, FLAME-conditioned deformation branch predicting per-Gaussian attribute residuals with spatial shape conditioning of the identity branch, disentangling identity from expression;
    \item An effective adversarial training strategy using a dual-discriminator on both rendered images and synthetic geometry cues to enforce expression consistency;
    \item State-of-the-art results in controllable animation, demonstrating superior expression fidelity and real-time performance (250 FPS on a single GPU, and $\sim$9 FPS on CPU) over previous animatable avatar methods.
\end{itemize}

\section{Related Work}
\label{sec:related_work}
Related works can be categorized into static 3D GANs and animatable 3D GANs.

\medskip\noindent
\textbf{Static 3D head avatars.}
A significant line of research focuses on training generative adversarial networks on 2D image datasets like FFHQ \cite{stylegan} to produce 3D-consistent outputs. Early works \cite{pix2nerf, pigan, gdna, graf, shadegan, gram, giraffe, stylesdf, cips3d, stylenerf} incorporated implicit neural radiance fields (NeRF) \cite{nerf} into 2D GAN frameworks to enable 3D-consistent generation. To improve rendering speed and view consistency, hybrid NeRF-based representations were introduced \cite{nerf_voxel, hexplane, instant_ngp} and adopted by subsequent methods \cite{eg3d, epigraf, voxgraf, volumegan, whatgan}. Several methods employ a super-resolution network to further accelerate rendering at the cost of some view consistency \cite{eg3d, volumegan, giraffe, stylesdf}. In particular, EG3D \cite{eg3d} builds upon the StyleGAN \cite{stylegan2} architecture to generate an intermediate triplane representation, which is then rendered into a low-resolution image and further upscaled. While producing high-fidelity results, this reliance on NeRF leads to slow training and inference and introduces 3D inconsistency artifacts due to the super-resolution network. To address this performance bottleneck, subsequent works replaced the NeRF renderer with the highly efficient 3D Gaussian Splatting (3DGS) representation. For instance, GGHEAD \cite{gghead} generates 3DGS attributes in a UV space mapped to a 3DMM template, while GSGAN \cite{gsgan} directly generates a 3D point cloud in a coarse-to-fine manner. Although these methods achieve real-time rendering, they are fundamentally designed for static head generation.

\medskip\noindent
\textbf{Dynamic 3D head avatars.} Another line of work extends 3D GANs to model facial dynamics \cite{DiscoFaceGAN, Next3D, anifacegan, eva3d, GNARF, fenerf, ide3d, OmniAvatar}. Some methods control articulations by conditioning generation on semantic maps \cite{fenerf, ide3d, eva3d}; however, such conditioning can yield inconsistent animation in videos. An alternative approach is to condition the generator on parameters of 3D morphable models (3DMM) \cite{FLAME, bfm_2009, SMPL}. Several approaches \cite{OmniAvatar, anifacegan, GNARF} also predict deformation fields to warp a canonical representation with respect to given 3DMM parameters. One recent work, Next3D \cite{Next3D}, incorporates 3DMM more directly: it adapts the EG3D framework by rasterizing neural textures on an articulated FLAME mesh and projecting the head onto orthogonal triplanes. To enforce geometric consistency, it introduces dual discrimination supervising the generator with synthetic renderings of the FLAME geometry. While this enables animation control, expression fidelity can degrade on complex motions, and dependence on an implicit representation leads to slow inference. Our work addresses this limitation by integrating animation control into a 3DGS-based 3D GAN, achieving both high-fidelity animation and real-time performance. Recent work GAIA \cite{gaia} explores a similar direction and delivers strong results. Compared with GAIA, AGORA applies spatial conditioning in both generator and discriminator (UV-aligned shape maps and FLAME-guided displacement renderings), which we find to be an important contributor to visual quality and expression fidelity (Sec.~\ref{subsec:ablation}).

\vspace{-.1cm}\medskip\noindent
\textbf{3D head avatars from single image.} Orthogonal to the generative task, another goal is to perform animation directly from a single input image \cite{headgan, zakharov_keypoints, dvp, face2face, NOFA}. Some recent methods, such as Live3DPortrait \cite{L3DP}, focus on generating high-quality static portraits, which, while impressive, do not produce animatable models. Building on this idea, methods like Portrait4D \cite{p4d, p4dv2} and VOODOO \cite{voodoo3d, voodooxp} aim to produce fully animatable avatars. These approaches typically operate by leveraging a pre-trained 3D GAN, either by distilling it into a direct image-to-avatar encoder or by training an encoder on a large-scale synthetic dataset generated by the GAN. While these models offer impressive single-shot performance, their quality is fundamentally dependent on the underlying generative model. A recent direction synthesizes pseudo-4D supervision with multi-view image/video diffusion models and then optimizes a personalized avatar (\eg, CAP4D and MVP4D \cite{cap4d,mvp4d}); however, this optimization-heavy process is time-consuming for each new subject. Another line uses strong DINO \cite{dinov2} features to infer animatable 3D Gaussian heads from a single image, including GAGAvatar \cite{gaga} and related methods \cite{perchead, flexavatar}. Although effective, these feature-driven approaches rely on explicit 3D supervision, whereas AGORA is trained purely from static 2D image collections. LAM \cite{lam} also conditions on DINO features, but its inference-time animation is based on plain linear blend skinning (LBS), restricting non-linear expression effects such as wrinkles and other fine-scale deformations. In contrast, our work strengthens the underlying 3D-aware GAN prior with an expression-specific deformation branch that captures such non-linear effects, providing a stronger foundation for single-image pipelines while retaining real-time inference.

\section{Method}
\label{sec:method}
\vspace{-.2cm}

In this section we introduce preliminary concepts followed by a concise description of our approach.

\subsection{Preliminaries}

\medskip\noindent
\textbf{3D Gaussian Splatting.} 3DGS \cite{3dgs} is an explicit 3D representation that models a scene as a collection of 3D Gaussians, enabling high-quality, real-time rendering. Each Gaussian is defined by a position $\mu \in \mathbb{R}^3$, a covariance matrix $\Sigma \in \mathbb{R}^{3\times3}$ (parameterized by a scale vector and a rotation quaternion), view-dependent color represented by spherical harmonics (SH) coefficients $c$, and an opacity $\alpha$. The color $C$ of a pixel is computed by alpha-blending $N$ Gaussians sorted by depth:
\begin{equation}
    C = \sum_{i=1}^{N} c_i \alpha'_i \prod_{j=1}^{i-1} (1 - \alpha'_j),
\end{equation}
where $\alpha'_i$ is the opacity of the $i$-th Gaussian modulated by its 2D projection onto the pixel. This process is differentiable, allowing for gradient-based optimization of all parameters.

\medskip\noindent
\textbf{FLAME 3D Morphable Model.} FLAME \cite{FLAME} is a statistical head model that provides a low-dimensional parametric space for shape $\beta$, expression $\psi$, and pose $\theta$. The template $T_P$ is formed by adding linear combinations of shape blendshapes $B_S(\beta)$, expression blendshapes $B_E(\psi)$, and pose-corrective blendshapes $B_P(\theta)$ to a base template $\bar{T}$:
\begin{equation}
T_P(\beta, \psi, \theta) = \bar{T} + B_S(\beta; \mathcal{S}) + B_E(\psi; \mathcal{E}) + B_P(\theta; \mathcal{P}).
\end{equation}

After applying linear blend skinning $W$, the articulated model is $
    M(\beta, \psi, \theta)=W(T_P(\beta,\psi,\theta), \mathcal{J}(\beta), \theta, \mathcal{W}).
$
This formulation allows for fine-grained control over facial identity and articulation. In this work, we primarily leverage the expression parameters $\psi$ and the jaw pose $\theta$ to drive the animation of our generated avatars.
\vspace{-.2cm}

\subsection{Architecture Overview}

\begin{figure}[t]
    \centering
    \includegraphics[width=0.4405\textwidth]{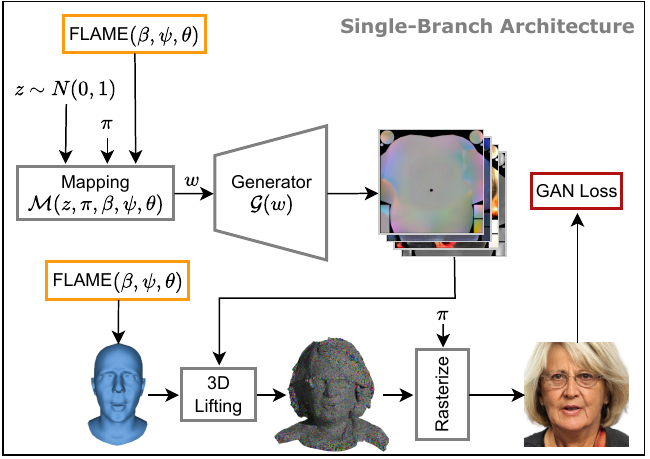}\hfill
    \includegraphics[width=0.5545\textwidth]{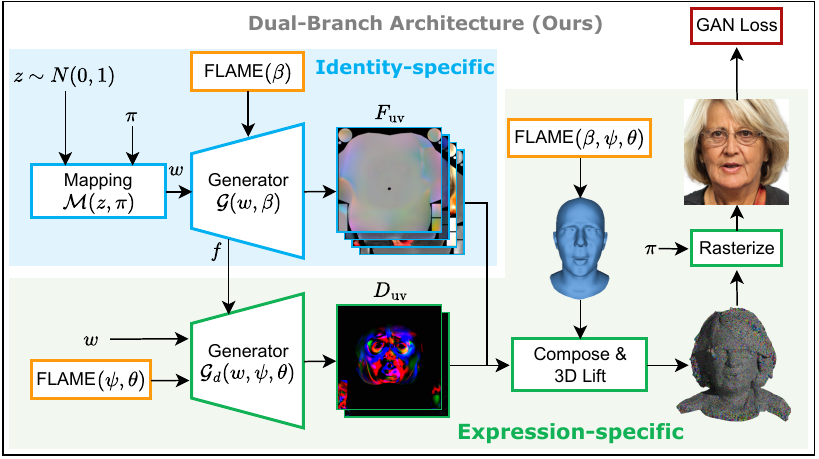}\vspace{-.15cm}\\
    {\small\makebox[0.4405\textwidth][c]{(a)}\hfill\makebox[0.5545\textwidth][c]{(b)}}\vspace{-.2cm}\\
    \caption{Single- and dual-branch generator designs and expression conditioning. (a) Single-branch GGHEAD-style baseline with no FLAME conditioning; the orange box marks the conditioning hook used in our ablations: $(\beta,\psi,\theta)$ injected into the mapping network $\mathcal{M}$. (b) Dual-branch AGORA: an identity path $\mathcal{G}(w,\beta)$ produces canonical 3DGS attributes, while an expression path $\mathcal{G}_{\text{d}}(w,\psi,\theta; f)$ predicts residuals. Residuals are composed with the canonical attributes, \emph{3D lifted} onto the articulated FLAME mesh, and rasterized.}
    \label{fig:overall_pipeline}
\end{figure}

The overall architecture of our generative model $\mathcal{G}$, illustrated in Fig.~\ref{fig:overall_pipeline}(b), builds upon the UV-based GGHEAD framework. Our goal is to produce an animatable 3DGS avatar from a latent code $z \in \mathcal{Z}$, camera parameters $\pi$, and FLAME parameters $(\beta, \psi, \theta)$. Following EG3D \cite{eg3d}, a mapping network $\mathcal{M}$ maps the latent code $z$ and camera parameters $\pi$ to an intermediate latent variable $w \in \mathcal{W}$. The core generator $\mathcal{G}$ then synthesizes a set of UV-space feature maps $F_\text{uv}$ from $w$ and $\beta$, which encode the identity-specific, canonical Gaussian attributes:
\begin{align}
w &= \mathcal{M}(z, \pi), \ \ F_\text{uv} = \mathcal{G}(w, \beta).
\end{align}
We obtain the raw Gaussian attributes $\mathcal{A}_{M}$ by bilinearly sampling $F_\text{uv}$ at a set of pre-defined UV coordinates $x_{uv}$:
\begin{align}
    \mathcal{A}_{M} = \text{GridSample}(F_\text{uv}, x_{uv}),
\end{align}
where $\mathcal{A}_{M} = \{\mu_{\delta M}, s, q, c, \alpha\}$ includes a position offset $\mu_{\delta M}$ relative to the template mesh $M$, as well as scale $s$, rotation $q$, color $c$, and opacity $\alpha$.  

The final 3D position of each Gaussian is obtained via \emph{3D lifting}: we bilinearly interpolate a base position from the articulated FLAME mesh $M(\beta,\psi,\theta)$ at $x_{uv}$ and add the predicted offset,
\begin{align}
    \mu = \text{Interpolate}(M(\beta,\psi,\theta), x_{uv}) + \mu_{\delta M}.
\end{align}
This design anchors the generated 3DGS to the underlying parametric mesh, providing a structured basis for animation. 
\vspace{-.4cm}

\subsection{Spatial Shape Conditioning}\vspace{-.1cm}
We condition the main generator $\mathcal{G}$ on the FLAME shape code $\beta$, as shown in Fig.~\ref{fig:overall_pipeline}(b), to encode shape-specific priors (\eg smaller craniofacial proportions for children). We found that naively injecting $\beta$ into the mapping network $\mathcal{M}$ makes the intermediate latent $w$ overly dominated by $\beta$, which reduces $z$-driven diversity and leads to mode collapse.

Instead, we adopt a softer, spatial conditioning strategy: we derive a UV-aligned map of the \emph{shape-isolated} deformation field and concatenate it to the feature maps within $\mathcal{G}$. Concretely, with $\beta_{0}$ denoting the canonical neutral shape code and neutral expression and jaw pose set to $(\psi_0,\theta_0)$, we compute
\begin{equation}
\Delta V_{\text{shape}}(\beta)
= M(\beta,\psi_0,\theta_0)
- M(\beta_{0},\psi_0,\theta_0).
\end{equation}
We then bake $\Delta V_{\text{shape}}$ into a UV-aligned displacement map, apply \emph{per-sample variance normalization} to better match the unit-variance assumption of the StyleGAN-style synthesis network,
and concatenate final channels with the block features. This spatial, UV-consistent conditioning injects shape biases where they matter geometrically while preserving the stochasticity carried by $z$.
\vspace{-.4cm}

\subsection{Deformation Branch}\vspace{-.1cm}
To refine the coarse 3DMM-based articulation and add high-frequency changes, we introduce a separate, lightweight deformation branch $\mathcal{G}_\text{d}$. This branch takes low-resolution $64\times64$ feature maps $f$ from the main generator, which encode coarse structural features (\eg, face shape, hair regions) \cite{stylegan}, and is conditioned on the FLAME expression $\psi$ and jaw pose $\theta$ via style modulation. It produces expression-specific features $D_\text{uv}$ for the geometric Gaussian attributes:
\begin{equation}
    D_\text{uv} = \mathcal{G}_\text{d}(w, \psi, \theta; f).
\end{equation}
From these, we obtain residual Gaussian attributes $A_D$ by bilinearly sampling $D_\text{uv}$ at UV coordinates $x_{uv}$, where $\mathcal{A}_{D} = \{\Delta\mu, \Delta s, \Delta q\}$.
These residuals are then \emph{composed} with the canonical attributes from the main branch to obtain the final Gaussian attributes $\mathcal{A}$. Specifically, we use post-activation summation for means, quaternion multiplication for rotations, and summation for log-scales.

This design is crucial for two reasons: (1) it prevents identity--expression entanglement, as naively conditioning the main generator $\mathcal{G}$ on expression parameters would cause the model to learn a unique identity for each expression vector; and (2) it enables fast, real-time animation of a fixed identity, since the more computationally expensive identity-specific $\mathcal{G}$ only needs to be run once.
\vspace{-.4cm}

\subsection{Final Gaussian Model}\vspace{-.1cm}
We apply activation functions similar to GGHEAD on the predicted position and scale of the Gaussians. For the position offsets, we use a bounded $\tanh$,
which upper-bounds the maximal deviation of Gaussians from the template mesh by $\gamma_\text{pos}$.  
For the scale parameters, we use a bounded exponential with softplus,
which constrains the maximum scale to $e^{-s_\text{max}}$ while initializing it at $e^{-s_\text{init}}$.  
Finally, we rasterize the Gaussian model $\mathcal{A}$ with camera parameters $\pi$ to produce the generated image.
\vspace{-.4cm}

\subsection{Enforcing Expression Consistency}\vspace{-.1cm}
\label{subsec:dual_discrimination}

\begin{figure}[t]
    \centering
    \includegraphics{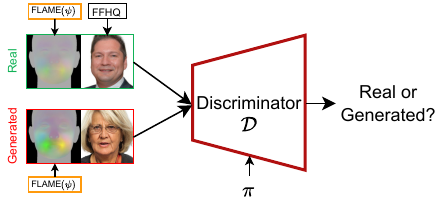}\vspace{-.2cm}
    \caption{Dual-Discrimination. We pass synthetic renderings $S(\psi)$ along with the RGB images.}
    \label{fig:dual_discrimination}
    \vspace{-.6cm}
\end{figure}

Naively training the generator with only an image-based discriminator is insufficient for precise expression control. Following Next3D \cite{Next3D}, we condition the discriminator $\mathcal{D}$ on the target expression by concatenating the rendered image with a synthetic FLAME rendering $\mathcal{S}(\psi)$, as illustrated in Fig.~\ref{fig:dual_discrimination}. With UV-coordinate coloring of FLAME vertices \cite{dvp}, expression consistency improves, but the model still under-expresses high-intensity cues (Sec.~\ref{subsec:ablation}).

We therefore replace UV coloring with a stronger displacement-based signal so that the discriminator can penalize fine-grained deviations. Specifically, we color-code the posed FLAME mesh by its expression-isolated vertex displacement from the neutral pose:
\begin{equation}
    \Delta V = V_{\text{posed}} - V_{\text{neutral}}.
\end{equation}
Using the same neutral-state notation as above, $V_{\text{posed}} \in M(\beta_{0}, \psi, \theta_0)$ and $V_{\text{neutral}} \in M(\beta_{0}, \psi_0, \theta_0)$. As indicated in Fig.~\ref{fig:dual_discrimination}, the resulting displacement-colored rendering $\mathcal{S}(\psi)$ is concatenated with the RGB render as input to $\mathcal{D}$. As shown in Table~\ref{tab:dd_ablation}, this significantly improves expression consistency.
\vspace{-.4cm}

\subsection{Loss functions \& Regularizations}
We train with the generator-side non-saturating GAN loss \cite{gan}:
\begin{equation}
    \mathcal{L}_\text{GAN}=\text{softplus}\big(-\mathcal{D}(\mathcal{R}(\mathcal{A}, \pi),\, \mathcal{S}(\psi),\, \pi)\big).
\end{equation}
Following GGHEAD, we also regularize position, scale, and opacity, employing $L_2$ losses $\mathcal{L}_\text{pos}$, $\mathcal{L}_\text{scale}$ and $\text{Beta}$ loss $\mathcal{L}_\text{opacity}$.
In a similar fashion, we apply $L_2$ regularization to the position $\mathcal{L}^{D}_\text{pos}$ and scale $\mathcal{L}^{D}_\text{scale}$ residuals predicted by the deformation branch.
These terms discourage large global warps and encourage the deformation branch to remain localized to fine-grained expression details.
Following prior works, we apply R1 regularization \cite{r1_gan,stylegan2} on Discriminator to keep training stable.
\vspace{-.4cm}

\section{Experiments}
\label{sec:experiments}

\begin{figure}[t]
    \centering
    \includegraphics[width=\linewidth]{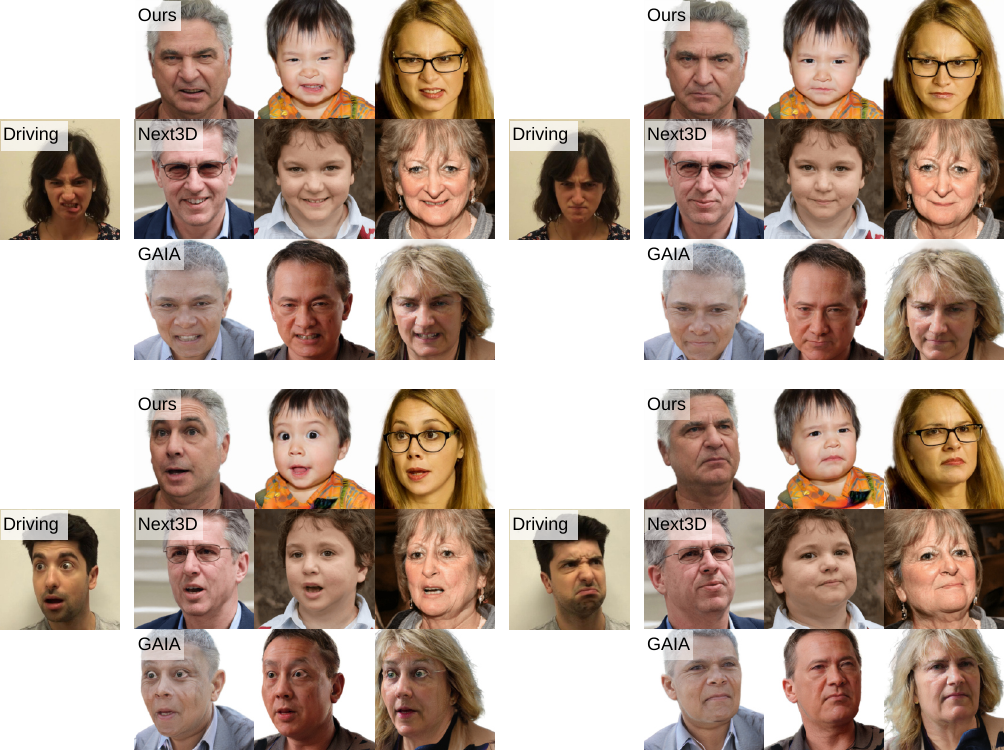}
    \caption{Qualitative comparisons with Next3D and GAIA on avatar generation, pose and face expression transfer.}
    \label{fig:generated_avatars}
    \vspace{-.6cm}
\end{figure}

\textbf{Implementation details.} 
Our generator and discriminator models are based on StyleGAN-2 architecture. We train the model in 2 stages. First, we train it for 6.5M images using $256\times256$ resolution rasterization, sampling 65K Gaussians on UV grid. Next, we train the model for 14M images on full $512\times512$ resolution, sampling 262K Gaussians. During both stages we use batch size 32, we set discriminator learning rate to $0.002$ and generator learning rate to $0.0025$. We apply R1 gradient penalty once every $16$ steps with $\gamma=1.0$ coefficient. For regularizations, we empirically choose $\lambda_{\text{pos}} = 0.25, \lambda_{\text{scale}}=0.5$ for main branch and $\lambda^{D}_{\text{pos}}=1.5, \lambda^{D}_{\text{scale}}=1.5$ for deformation branch. We apply $\mathcal{L}_{\text{opacity}}$ only during the second stage and use coefficient $\lambda_\text{opacity}=1.0$. The entire training takes 4 days on 4$\times\text{RTX A6000}$.   

\smallskip\noindent
\textbf{Dataset.} We train our model on FFHQ \cite{stylegan} dataset, consisting of 70000 human head images. Following prior works, we mirror the dataset to obtain around 140000 images in total. Following GGHEAD~\cite{gghead}, we use MODNet~\cite{MODNet} to remove background from FFHQ images. For each image we estimate FLAME parameters via off-the-shelf estimator~\cite{smirk}, and further derive camera parameters from estimated head rotations. 

\smallskip\noindent
\textbf{Baselines.} 
We compare with the static baselines GGHEAD~\cite{gghead} and EG3D~\cite{eg3d}. We also compare with the state-of-the-art NeRF-based animatable 3D GAN, Next3D~\cite{Next3D}, and the recent 3DGS-based method GAIA~\cite{gaia}.

We employ Fr\'echet Inception Distance (FID) to measure the quality of generated images. Following prior works \cite{Next3D}, we use Average Pose Distance (APD), Average Expression Distance (AED) and identity consistency (ID)  metrics to evaluate animation quality. To further evaluate mouth consistency, similar to GAIA we calculate the Average Jawpose Distance (AED-jaw).  To compute metrics, we follow exactly the same protocol described in \cite{Next3D}.

We further compare inference speed of the methods on avatar reenactment setting, \ie with running identity branch once with caching. To this end, we report FPS on a single RTX A6000 of PyTorch implementation of our method without any additional optimizations. For Next3D we measure FPS on the same machine using the code from the author's repository \cite{Next3D}. %
For GAIA we take the measurements from their paper, which are also conducted on an RTX A6000 GPU with identity caching. In addition, we report CPU-only inference speed for AGORA; hardware details are given in the supplementary material.
\vspace{-.5cm}

\begin{figure}[t]
    \centering
    \begin{minipage}[t]{0.49\linewidth}
        \centering
        \includegraphics[width=\linewidth]{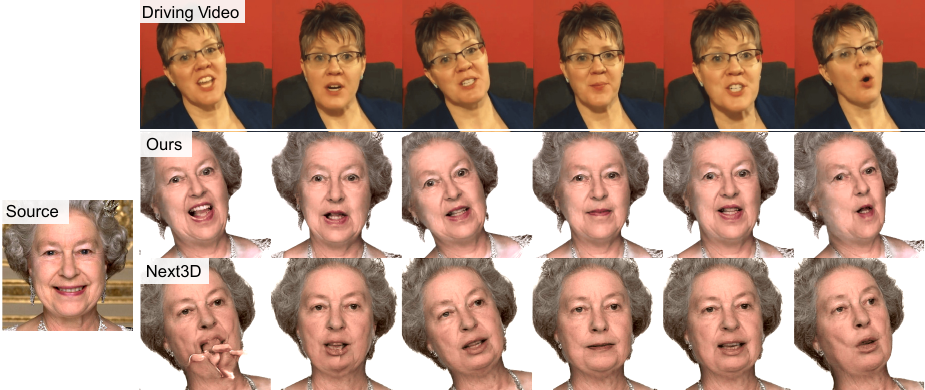}
        {\footnotesize (a) Next3D}
    \end{minipage}\hfill
    \begin{minipage}[t]{0.49\linewidth}
        \centering
        \includegraphics[width=\linewidth]{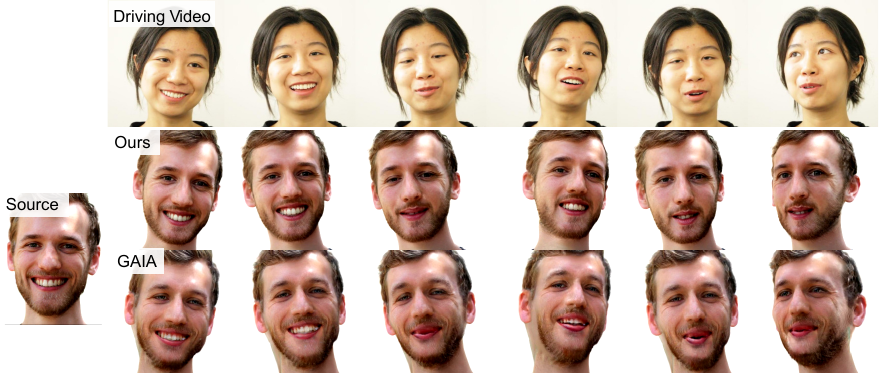}
        {\footnotesize (b) GAIA}
    \end{minipage}
    \caption{Single-image avatar (PTI) comparisons against previous methods.}
    \label{fig:pti_avatars_next3d}
    \label{fig:pti_avatars_gaia}
    \vspace{-.2cm}
\end{figure}

\subsection{Comparison with SOTA}
Table~\ref{table:main_metrics} compares AGORA with static head generators GGHEAD~\cite{gghead}, EG3D~\cite{eg3d} and recent animatable head avatars Next3D~\cite{Next3D}, GAIA~\cite{gaia}. As can be seen, our approach outperforms all other baselines in terms of image quality (FID).
For camera and identity consistency, AGORA scores better in APD and ID compared to GAIA and Next3D. For the animation quality, we significantly outperform NeRF-based Next3D in both AED and AED-jaw. Compared to GAIA, our method produces more consistent mouth articulations, resulting in significantly better AED-jaw. While GAIA shows better expression consistency in AED metrics, qualitative comparison in Fig.~\ref{fig:pti_avatars_gaia} demonstrates fewer artifacts in faces produced by our method. Finally, our method significantly outperforms previous methods in inference speed. We report 250 FPS and $\sim$9 FPS runtime on GPU and CPU (16 threads), respectively, enabled by the lightweight deformation branch ${\cal{G}}_d$ and the simple FLAME linear blend skinning used by AGORA.

\begin{table}[t]
\centering
\caption{
Comparison with state of the art. Lower values are better for FID/AED/APD; higher is better for ID/FPS. Best results are shown in bold, and second-best are underlined. Note that EG3D and GGHEAD don't offer expression control, so we only report FID.
}
\vspace{-.3cm}
\setlength{\tabcolsep}{4.85pt}      %

{\small  
    \begingroup
    \setlength{\tabcolsep}{0.07cm}
    \begin{tabular}{lcccccr}
        \toprule
        Method & FID $\downarrow$ & $\text{AED}$ $\downarrow$ & $\text{AED-jaw}$ $\downarrow$ & $\text{ID}\uparrow$ & $\text{APD}$ $\downarrow$ & $\text{FPS}$ $\uparrow$ \\
        \midrule
        EG3D~\cite{eg3d} & 3.28 & -- & -- & -- & -- & -- \\
        GGHEAD~\cite{gghead} & 4.06 & -- & -- & -- & -- & -- \\
        Next3D~\cite{Next3D} & \underline{3.18} & 0.930 & 0.046 & \underline{0.74} & 0.031 & 15 \\
        GAIA~\cite{gaia} & 3.85 & \textbf{0.530} & \underline{0.040} & 0.72 & \underline{0.027} & \underline{52} \\
        AGORA & \textbf{3.17} & \underline{0.682} & \textbf{0.021} & \textbf{0.75} & \textbf{0.025} & \textbf{250} \\
        \bottomrule
    \end{tabular}
    \vspace{-0.6cm}
    \endgroup
}
\label{table:main_metrics}
\end{table}

\subsection{Qualitative Results}

\textbf{Avatar generation.}
We qualitatively compare AGORA (ours) against Next3D and GAIA on avatar expressiveness. Fig.~\ref{fig:generated_avatars} presents results of re-animating faces with strong expressions from the FEED dataset \cite{drobyshev2024emoportraits}. Our method closely follows the driving expressions and recovers fine-grained cues such as wrinkles, while remaining emotion-consistent (\ie clear \emph{disgust}, \emph{anger}, and \emph{surprise}). Notably, for the same \emph{surprise} input our model yields age-appropriate behavior (forehead wrinkling for older subjects but not for children), enabled by shape-code conditioning, which preserves identity priors while modulating expression detail. While Next3D renders high-quality images, it often under-expresses or misaligns the target articulation in high-intensity cases. GAIA generally produces valid expressions, but AGORA renders more realistic faces and shows fewer artifacts in extreme expressions, where GAIA occasionally exhibits issues around the eyes.

\smallskip\noindent
\textbf{Avatars from a single image.}
We further compare methods by generating avatars from a single image using Pivotal Tuning Inversion (PTI)~\cite{pti}. For each source face image we run PTI to obtain its avatar, randomly assign a driving video, and transfer its expressions and camera motion. For both AGORA and Next3D we use the same PTI implementation from the authors' repository \cite{pti}. For GAIA, neither code nor model weights are publicly available, so we report only the PTI example provided by the authors and exclude GAIA from the user study below, which requires personalizing and re-animating all source images.

\begin{figure}[t]
    \centering
    \includegraphics[width=0.55\linewidth]{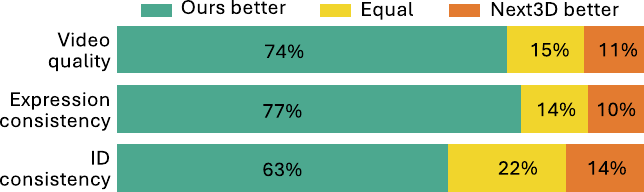}
    \caption{User study on single-image avatar reenactment. Pairwise preferences between AGORA and Next3D over identity consistency, expression consistency, and overall video quality.}
    \label{fig:user_study}
\end{figure}

As illustrated in Fig.~\ref{fig:pti_avatars_next3d} (left panel), Next3D struggles under extreme jaw poses, exhibiting pronounced mouth/teeth artifacts, whereas AGORA maintains precise mouth articulation and better preserves identity. To evaluate this trend at scale, we run a video-level user study with 50 in-the-wild source images and 20 driving videos.
The driving set includes challenging clips with large rotations and strong expressions from FEED~\cite{drobyshev2024emoportraits} and additional in-the-wild talking-head videos.
Participants compare AGORA and Next3D on identity consistency, expression consistency, and overall video quality (temporal smoothness).
Fig.~\ref{fig:user_study} shows consistent preference for AGORA across all three criteria; full protocol details are provided in the supplementary material.

Fig.~\ref{fig:pti_avatars_gaia} (right panel) presents a qualitative comparison of our method with GAIA. We can observe frequent artifacts in mouth areas and inconsistent head geometry for GAIA-generated avatars, see \eg, column 4. In contrast, our method yields smoother motion and consistent teeth rendering. These observations can be further confirmed from videos in the supplementary material.

\begin{wrapfigure}{r}{0.45\textwidth}
    \centering
    \vspace{-10pt}
    \includegraphics[width=0.96\linewidth]{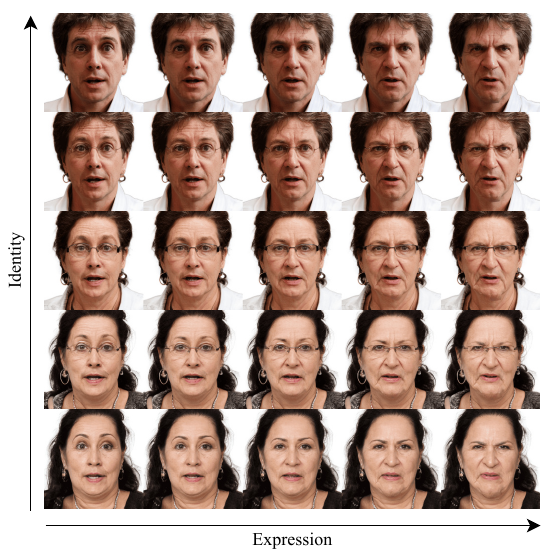}
    \caption{Expression--identity disentanglement. Rows vary expression at fixed identity; columns vary identity at fixed expression.}
    \label{fig:id_exp_disentanglement}
    \vspace{-12pt}
\end{wrapfigure}

\smallskip\noindent
\textbf{Expression--identity disentanglement.}
In Fig.~\ref{fig:id_exp_disentanglement} we assess disentanglement of expression and identity by independently interpolating the identity latent and the FLAME expression parameters. We sample two identities and two expressions and obtain other identity codes and expression settings by linear interpolation. 
We then render the resulting grid of expressions and identities while keeping other parameters unchanged. Rows exhibit smooth expression transitions with stable identity, while columns vary identity without altering the intended expression. The grid indicates clean factor separation: precise expression changes and consistent identity traits, with no visible entanglement.

\subsection{Ablation Study}
\label{subsec:ablation}

We next evaluate design choices of our method.
To keep ablations tractable, we report results for a lightweight setting using $256{\times}256$ image resolution and 65K Gaussians. Unless stated otherwise, all variants share the same data, schedule, and seeds under these settings. 
The exception is Table~\ref{tab:gaia_design_ablation}, which reports full-scale runs with $512{\times}512$ rasterization and 262K Gaussians.

\smallskip\noindent

\textbf{Conditioning architecture.} Table~\ref{table:ablations_eid} compares architectures for identity-consistent expression control. We ablate different ways to condition the model on FLAME shape, expression, and jaw parameters $(\beta,\psi,\theta)$.

\noindent Exp.1--2 are single-branch baselines following Fig.~\ref{fig:overall_pipeline}(a): Exp.1 uses only FLAME LBS and gives the worst FID/AED, showing explicit conditioning is necessary. Its high ID reflects under-articulation, not disentanglement: appearance changes little beyond the LBS geometry, so identity is easy to preserve.
Exp.2 injects $(\beta,\psi,\theta)$ directly into the mapping network $\mathcal{M}$ (orange hook in Fig.~\ref{fig:overall_pipeline}(a)); FID improves but ID drops, indicating identity--expression entanglement (visualized in Fig.~3 of the supplementary material).
Replacing that injection with the deformation branch (Exp.3) improves FID/AED further, yet ID remains low.
Exp.4 additionally conditions $\mathcal{G}$ on person-specific, expression-independent shape $\beta$, recovering ID at a small AED cost ($0.563\rightarrow0.588$) and giving the best overall FID/AED/ID trade-off. AED-jaw and APD stay essentially constant across all four variants.
Overall, splitting identity and expression pathways improves quality while enabling fast animation, since only lightweight expression-specific modules run at test time.

\begin{table}[t]
    \centering
    \caption{Expression--identity disentanglement. Lower is better for FID, AED, AED-jaw, and APD; higher is better for ID. ``S'' and ``D'' stand for Single and Dual branch architectures.}
    \vspace{-.1cm}
    \footnotesize
    \setlength{\tabcolsep}{4pt}
    \renewcommand{\arraystretch}{1.1}
    
    \setlength{\tabcolsep}{3pt}      %
    {\footnotesize  
        \begin{tabular}{lccc|ccccc}
            \toprule
            Exp. & $\cal{M}$ & $\cal{G}$ & Branch
            & \multicolumn{1}{c}{\makecell{FID$\downarrow$}}
            & \multicolumn{1}{c}{\makecell{AED$\downarrow$}}
            & \multicolumn{1}{c}{\makecell{AED-jaw$\downarrow$}}
            & \multicolumn{1}{c}{\makecell{ID$\uparrow$}}
            & \multicolumn{1}{c}{\makecell{APD$\downarrow$}} \\
            \midrule
            1. & - & - & S & 6.59 & 0.686 & \textbf{0.022} & \textbf{0.74} & \textbf{0.026} \\
            2. & $\beta,\psi,\theta$ & - & S & 5.46 & 0.664 & \textbf{0.022} & 0.56 & \underline{0.027} \\
            3. & - & - & D & \underline{5.29} & \textbf{0.563} & \underline{0.024} & 0.56 & \textbf{0.026} \\
            4. (Ours) & - & $\beta$ & D & \textbf{4.72} & \underline{0.588} & \textbf{0.022} & \underline{0.70} & \textbf{0.026} \\
            \bottomrule
        \end{tabular}
        \vspace{-.1cm}
    }
    \label{table:ablations_eid}
\end{table}

\smallskip\noindent

\textbf{Discriminator expression conditioning.} Table~\ref{tab:dd_ablation} compares discriminator conditioning strategies for expression control. Without expression-aware discrimination (row~1), the generator is free to optimize realism alone and produces much worse AED. Conditioning with cGAN-style vector projection \cite{cgan_projection} also fails to improve AED (row~2), likely because stronger camera/geometry cues dominate and the expression signal is ignored. We therefore condition through image space by concatenating synthetic renderings with real/generated images. Next3D-style dual discrimination gives only a minor AED gain (row~3): its renderings also encode FLAME's shape code, so the discriminator can memorize identity from each (shape, expression) pair, pushing the generator toward identity--expression entanglement (ID 0.59 vs. 0.70 for ours) rather than faithful expressions. Our expression-only rendering with LBS-displacement texture cues removes this shape leakage, provides the strongest expression geometry signal, and achieves the best AED/AED-jaw (row~4).

\begin{table}[t]
  \centering
  \caption{Dual-Discrimination ablation. Lower is better.}
    \vspace{-.1cm}
  \label{tab:dd_ablation}

        \begin{tabular}{lcc}
            \toprule
            Variant
            & \multicolumn{1}{c}{\makecell{AED$\downarrow$}}
            & \multicolumn{1}{c}{\makecell{AED-jaw$\downarrow$}} \\
            \midrule
            1. w/o dual discrimination & 0.832 & 0.024 \\
            2. cGAN-style vector projection & 0.847 & 0.025 \\
            3. Next3D-style dual discrimination& 0.766 & 0.024 \\
            4. Ours & \textbf{0.588} & \textbf{0.022} \\
            \bottomrule
        \end{tabular}
    \vspace{-.2cm}
\end{table}

\smallskip\noindent
\textbf{Comparison with GAIA design choices.}
In Table~\ref{tab:gaia_design_ablation} we report an additional ablation of our full-scale setup ($512{\times}512$, 262K Gaussians) aimed at comparing our method with GAIA.
We modify the following components of AGORA: spatial conditioning is replaced with GAIA's vector conditioning, and dual discrimination is replaced with GAIA's two-discriminator design (shape and expression).
GAIA* is our closest GAIA-style setting: both AGORA components are disabled, and we also apply GAIA's loss scheduling and deformation-branch regularization from their paper (official code was unavailable).
Every modified variant has higher FID than Ours (3.42--8.53 vs. 3.17).
Notably, spatial conditioning on its own is worse than using neither component (8.53 vs. 6.08), suggesting that shape-conditioned features only pay off once the discriminator is expression-aware, which is what dual discrimination provides.
The two components are therefore complementary, and together they account for most of the visual-quality gap to the GAIA-style baseline.
FID, however, does not capture identity preservation, where spatial conditioning has its clearest effect (Table~\ref{table:ablations_eid}, Exp.3 vs. Exp.4: ID $0.56\rightarrow0.70$).

\begin{table}[t]
  \centering
  \caption{Comparison with GAIA design choices in full-scale training ($512{\times}512$, 262K Gaussians). Lower is better for FID.}
  \vspace{-.1cm}
  \label{tab:gaia_design_ablation}
  \setlength{\tabcolsep}{4.2pt}
  {\small
    \begin{tabular}{lcc|c}
      \toprule
      Variant
      & \multicolumn{1}{c}{\makecell{Sp.\\cond.}}
      & \multicolumn{1}{c}{\makecell{Dual-\\disc.}}
      & \multicolumn{1}{c}{\makecell{FID$\downarrow$}}
       \\
      \midrule
      GAIA* & $\times$ & $\times$ & 6.08 \\
      Dual-disc. only & $\times$ & $\checkmark$ & 3.42 \\
      Sp. cond. only & $\checkmark$ & $\times$ & 8.53 \\
      Ours & $\checkmark$ & $\checkmark$ & \textbf{3.17} \\
      \bottomrule
    \end{tabular}
  }
  \vspace{-.2cm}
\end{table}

\section{Conclusion}

We presented AGORA, a conditional 3DGS GAN for animatable head avatars. A dual-branch generator couples an identity path with an expression-specific deformation branch; spatial shape conditioning injects shape priors without collapsing diversity; and dual discrimination on synthetic geometry cues enforces precise expressions. The same model applies to single-image PTI avatars and remains stable under large poses and articulations. AGORA results in excellent generation quality, real-time rendering at 250 FPS on a single GPU, and interactive-rate CPU rendering, while outperforming existing animatable avatar methods. Future work will address hair animation, robustness to varying illumination, as well as extensions to full-body models.

\bibliographystyle{splncs04}
\bibliography{main}

\clearpage
\appendix
\setcounter{figure}{0}
\setcounter{table}{0}
\renewcommand{\thefigure}{S\arabic{figure}}
\renewcommand{\thetable}{S\arabic{table}}
\renewcommand{\theHsection}{\Alph{section}}
\renewcommand{\theHfigure}{S\arabic{figure}}
\renewcommand{\theHtable}{S\arabic{table}}

\begin{center}
{\Large\bfseries\boldmath Supplementary Material\par}
\end{center}
\suppressfloats[t]
\vspace{1em}

This supplementary material provides additional results and more details about our method. In particular, Section~\ref{sec:supp_results} presents user study details and additional qualitative results. Section~\ref{sec:supp_method_details} provides template, architecture, and evaluation details. Finally, Section~\ref{sec:supp_discussion} discusses some of our design choices, limitations, and ethical considerations. Please refer to the supplementary video for more qualitative results.

\section{Additional Results}
\label{sec:supp_results}

\subsection{User Study}
This subsection describes the user-study protocol used for the comparison reported in the main paper. We run a video-level, three-question forced-choice study with an optional \texttt{equal} (tie) option. Each participant answers three questions per video: ID consistency, expression consistency, and overall video quality. Each participant rates 25 of the 50 video pairs; participants are split into two groups to cover all pairs. In the interface, we first show a static \emph{target} portrait (identity to match). Below it, each trial presents a three-panel comparison video: \emph{LEFT} = Method A, \emph{MIDDLE} = real driving video, \emph{RIGHT} = Method B. We randomly swap the left/right assignment of methods on every trial. In the UI, \texttt{method\_1} corresponds to ours and \texttt{method\_2} to Next3D.

\begin{figure}[t]
    \centering
    \includegraphics[width=\linewidth]{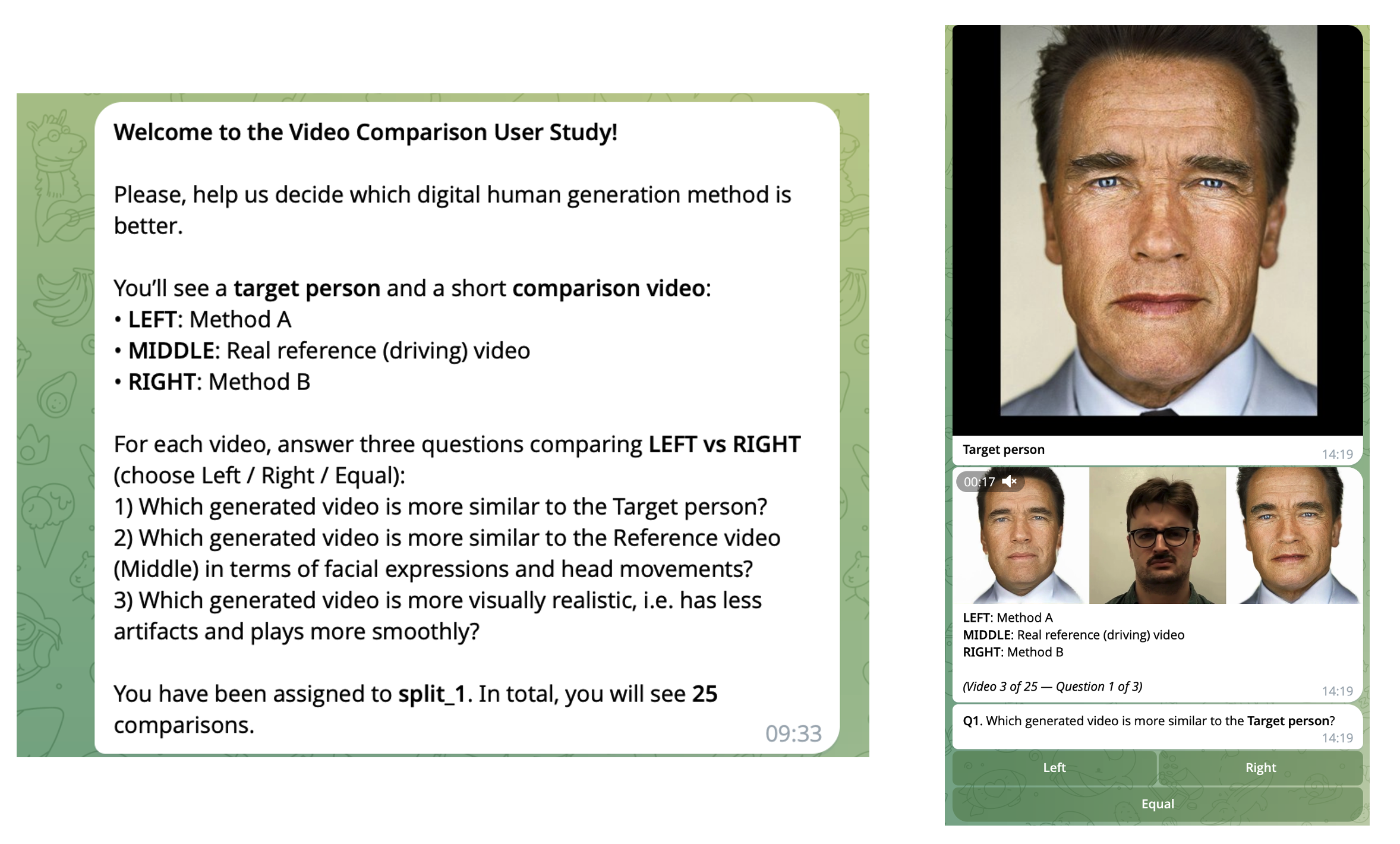}
    \caption{User-study interface.}
    \label{fig:user_study_ui}
\end{figure}

Fig.~\ref{fig:user_study_ui} shows the user-study interface, which we use for the AGORA comparison reported in Fig. 6 of the main paper.

We discard \texttt{equal} votes and evaluate significance with a two-sided binomial test against 50\% on decisive votes. Table~\ref{tab:user_study} summarizes the results. For all three questions, the preference for our method is highly statistically significant.

\begin{table}[t]
    \centering
    \caption{Pairwise preference (ours vs. Next3D) on decisive votes. We report two-sided binomial $p$-values.}
    \label{tab:user_study}
    \footnotesize
    \setlength{\tabcolsep}{4pt}
    \renewcommand{\arraystretch}{0.9}
    \setlength{\tabcolsep}{3pt}
    {\footnotesize
        \begin{tabular}{lccccc}
            \toprule
            & \multicolumn{1}{c}{\makecell{method\_1\\(ours)}}
            & \multicolumn{1}{c}{\makecell{method\_2\\(Next3D)}}
            & \multicolumn{1}{c}{\makecell{Ties\\(abs)}}
            & \multicolumn{1}{c}{\makecell{Win rate\\(ours, \%)}}
            & \multicolumn{1}{c}{\makecell{$p$-value\\(binomial)}} \\
            \midrule
            q1 (ID)         & 335 & 76  & 117 & 81.5 & $<10^{-39}$ \\
            q2 (Expression) & 405 & 51  & 72  & 88.8 & $<10^{-68}$ \\
            q3 (Quality)    & 389 & 58  & 81  & 87.0 & $<10^{-60}$ \\
            \bottomrule
        \end{tabular}
    }
\end{table}

\subsection{Additional Qualitative Results}

\subsubsection{Few-shot Inference.}
\begin{figure}[t]
    \centering
    \includegraphics[width=0.88\linewidth]{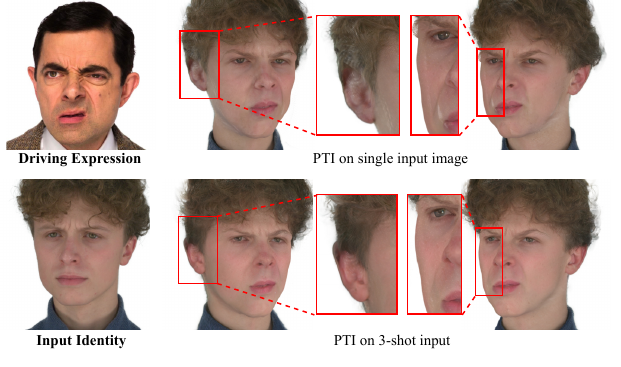}
    \caption{Few-shot PTI illustration. Top: PTI from a single input image. Bottom: PTI from three input images. Multi-frame tuning reduces view-dependent artifacts and yields more stable geometry.}
    \label{fig:fewshot_inference}
\end{figure}

In the main paper, we apply PTI only to single-image personalization (Sec. 4.2 and Fig. 5), but the same pipeline can also be extended naturally to a few-shot setting.
Fig.~\ref{fig:fewshot_inference} illustrates a simple extension of PTI from one-shot to few-shot personalization. With a single input image, PTI can overfit to the observed view and produce side-view artifacts on out-of-distribution identities. Using several input frames stabilizes the optimization and yields a more coherent avatar under viewpoint changes. In our few-shot setup, we jointly tune the pivot latent $z$ and the FLAME parameters $\beta, \psi, \theta$ across the available frames during the first PTI stage before continuing with the standard generator tuning.

\subsubsection{$z$-driven Diversity.}
\begin{figure}[t]
    \centering
    \includegraphics[width=\linewidth]{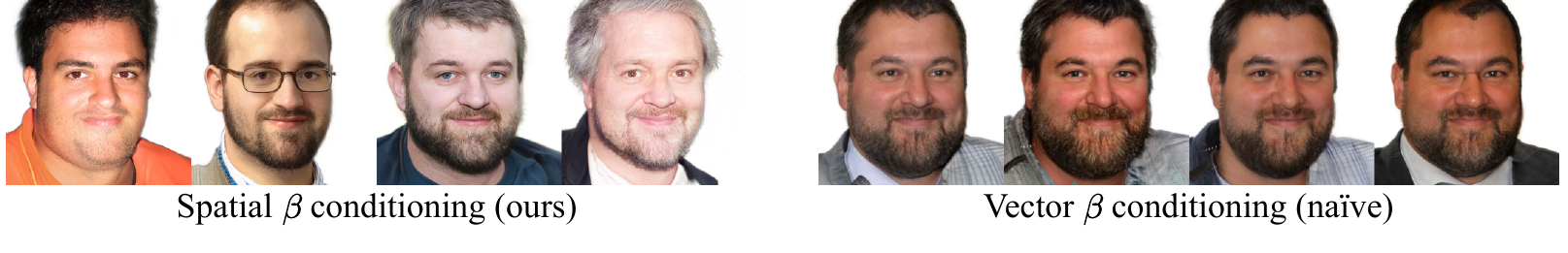}
    \caption{Qualitative comparison of latent-driven identity diversity. For each group, the driving parameters are fixed and only the latent code varies across columns.}
    \label{fig:z_driven_diversity}
\end{figure}

As discussed in Sec. 3.3 of the main paper, naive vector conditioning of the shape code can reduce identity diversity and lead to identity--expression entanglement; this is also reflected by the low ID score in Table 2, row 2 of the main paper. Here we visualize this failure mode by comparing naive vector conditioning against our spatial shape conditioning.
Fig.~\ref{fig:z_driven_diversity} demonstrates that AGORA preserves latent-driven identity diversity even when the driving tuple is fixed. We keep the FLAME shape, expression, pose, and camera parameters constant and vary only the latent code $z$ across columns. With our spatial conditioning, changing $z$ produces distinct identities while maintaining the same articulation. In contrast, a naive vector conditioning of the shape code collapses to nearly the same identity, which is consistent with the identity--expression entanglement observed in the ablation study.

\section{Method Details}
\label{sec:supp_method_details}

\subsection{Template and Architecture Details}

\subsubsection{FLAME Template with Mouth.}
We augment the standard FLAME2020 template with mouth interior because vanilla FLAME does not model visible structures such as the oral cavity and teeth, which leads to unrealistic mouth appearance under strong expressions and large jaw motion.
We extend FLAME2020 with a mouth cavity and two (upper/lower) rows of frontal teeth. For the mouth cavity, we derive skinning weights by averaging the upper- and lower-lip weights and mixing them 50/50 as vertices approach the deepest point of the cavity. For shape and expression blendshapes, we reuse those of the upper/lower lips. For pose blendshapes, we take the upper/lower-lip correctives and blend them 50/50 toward the cavity apex. Since our training data lacks back-head supervision, we remove the back-head region and fill the resulting UV hole by stretching neighboring side regions.

\subsubsection{Deformation Branch Details.}
\begin{figure}[t]
    \centering
    \includegraphics[width=0.46\textwidth]{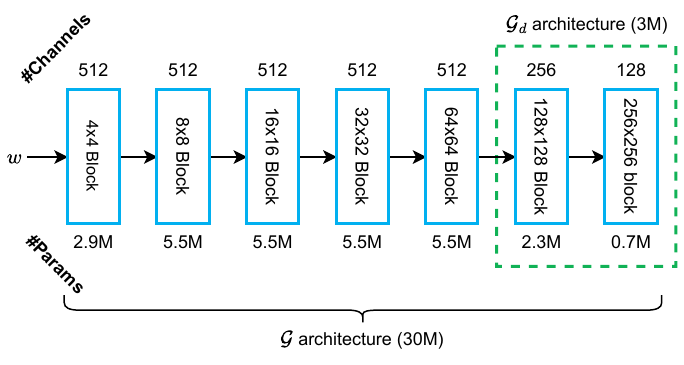}
    \caption{Architecture of identity- and expression-specific branches (StyleGAN2 blocks).}
    \label{fig:supp_detailed_architecture}
\end{figure}

We deliberately design the expression-specific branch to be lightweight. It takes as input $64\times64$ features from the main branch, which encode identity-related structure, and predicts $256\times256$ expression-specific UV-space deformations of the Gaussian attributes via two consecutive StyleGAN2 blocks (see Fig.~\ref{fig:supp_detailed_architecture}). The deformation branch has only $\sim$3M parameters, compared to $\sim$30M for the identity branch, enabling real-time avatar animation at inference.

\subsubsection{Spatial Shape Conditioning Details.}
\begin{figure}[t]
    \centering
    \includegraphics[width=0.45\textwidth]{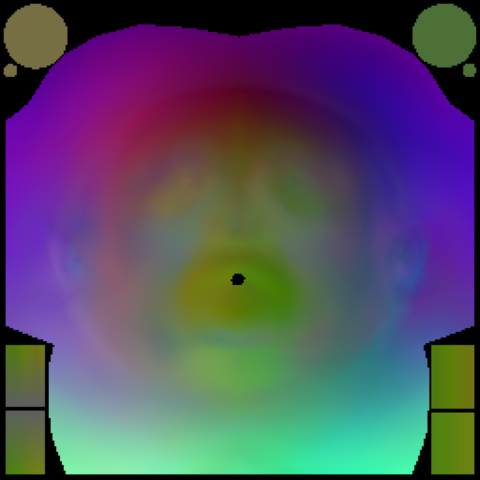}
    \caption{Each pixel encodes the vertex XYZ displacement w.r.t. the template.}
    \label{fig:supp_shape_cond}
\end{figure}

Given the FLAME shape code, we generate a UV-aligned field of shape-induced XYZ residuals (see Fig.~\ref{fig:supp_shape_cond}) and inject these features into every StyleGAN block of the main generator.

\subsection{Evaluation Details}

\subsubsection{Metric Computation.}
We follow the evaluation protocol of Next3D~\cite{Next3D}.  
FID is computed on FFHQ using 50k random samples drawn from the latent, camera, and FLAME parameter distributions; for $256^2$ baselines we use FFHQ-256.  
For AED and APD, we randomly sample 500 identities and, for each, 20 random \{expression, pose\} pairs. We re-estimate FLAME parameters from generated images and compute the mean distance to the driving parameters. 
The ID score is calculated as the mean ArcFace cosine similarity between 1000 generated image pairs of the same identity under different poses and expressions. Cropping and alignment are consistent with training.

\subsubsection{Hardware Details.}
Our GPU runtimes are measured on a single NVIDIA RTX A6000, using a naive PyTorch implementation of our method without any additional inference optimizations. For CPU inference measurements, we use an Intel Xeon Platinum 8570 CPU and run naive PyTorch CPU inference with 16 threads. Following the reenactment protocol of the main paper, the identity branch is evaluated once per subject and cached, so the reported frame rates correspond to running the deformation branch, attribute composition, 3D lifting, and rasterization per frame.

\section{Discussion}
\label{sec:supp_discussion}

\subsection{Design Choices}

\subsubsection{Comparison to Next3D's Dual Discrimination.}
\begin{figure}[t!]
    \centering
    \includegraphics[width=0.46\textwidth]{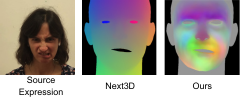}
    \caption{Comparison of synthetic renderings $S(\psi)$.}
    \label{fig:supp_dd_comparison}
\end{figure}

We visually compare our dual-discrimination synthetic renderings with those from the Next3D approach (Fig.~\ref{fig:supp_dd_comparison}). We observe that our renderings provide stronger expression-specific cues, enabling the discriminator to more effectively penalize the generator for expression inaccuracies.

\subsection{Discussion and Limitations}
\noindent
\textbf{Why not diffusion?}
Our goal is \emph{real-time} sampling and reenactment with explicit FLAME control and identity caching; GANs provide single-pass generation and fit naturally with our dual-branch formulation.
A diffusion formulation would require a well-defined denoising target in 3D-representation space (\eg, UV-parameterized 3DGS attribute maps), which is not available as ground-truth for in-the-wild 2D datasets like FFHQ; alternatively, latent diffusion would require an additional image$\rightarrow$3DGS encoder/autoencoder pipeline to define the latent space. This direction is orthogonal to the scope here.

\smallskip\noindent
\textbf{Use of larger datasets.}
The current StyleGAN2-based training recipe remains stable at the scale used in this paper, aided by R1 and Gaussian regularization. Scaling to substantially larger datasets is feasible in principle, but the benefit will likely depend more on the diversity of head pose and facial expression than on raw image count alone. Extending the data distribution along these axes is therefore a more promising direction than merely increasing volume.

\smallskip\noindent
\textbf{FLAME expressivity.}
AGORA inherits the expressive ceiling of FLAME, especially for highly asymmetric or otherwise out-of-space facial motions. The framework itself is modular: it can be retrained with a more expressive FLAME-like model, and it can also be extended with learned implicit expression latents that complement the explicit FLAME controls at the cost of additional inference-time computation.

Beyond these directions, our current model is trained and evaluated on frontal heads and does not synthesize the back of the head. This limitation can be alleviated by supervising with full-head data, in the spirit of PanoHead~\cite{panohead}. In addition, we do not explicitly control ocular gaze: the generator is not conditioned on gaze and the FLAME parameters we estimate exclude eyeball rotations. A practical path forward is to obtain per-image gaze from a monocular estimator during training and map these predictions to FLAME's eyeball joints, enabling explicit gaze control at inference. We also leave hair animation and illumination handling to future work.

\subsection{Ethical Considerations}
AGORA produces identity-preserving, controllable head avatars at real-time rates, even on CPU, which lowers the barrier to large-scale deployment and introduces dual-use risks (\eg, unauthorized identity cloning, consentless personalization, and misuse in telepresence or misinformation). In our experiments we rely on publicly available datasets and do not train on private images; for any release we will provide usage guidelines that require proof of consent for single-image personalization. We explicitly discourage use for impersonation and encourage research on consent-aware editing and robust content provenance to complement technical advances.

\end{document}